\documentclass[sigconf, nonacm]{acmart}

\makeatletter
\def\@ACM@checkaffil{
    \if@ACM@instpresent\else
    \ClassWarningNoLine{\@classname}{No institution present for an affiliation}%
    \fi
    \if@ACM@citypresent\else
    \ClassWarningNoLine{\@classname}{No city present for an affiliation}%
    \fi
    \if@ACM@countrypresent\else
        \ClassWarningNoLine{\@classname}{No country present for an affiliation}%
    \fi
}
\makeatother

\newcommand{\one}{\textbf{({\em i}\/)}\xspace}
\newcommand{\two}{\textbf{({\em ii}\/)}\xspace}
\newcommand{\three}{\textbf{({\em iii}\/)}\xspace}
\newcommand{\four}{\textbf{({\em iv}\/)}\xspace}
\newcommand{\five}{\textbf{({\em v}\/)}\xspace}

\def\eg{\emph{e.g., }\xspace}

\def\ie{\emph{i.e. }\xspace}
\def\etal{\emph{et al.}\xspace}
\def\vs{\emph{vs.}\xspace}

\newcommand{\pb}[1]{\vspace{0.75ex}\noindent{\bf \em #1}\hspace*{.3em}}
\usepackage{xspace}
\usepackage{csquotes}
\usepackage{subcaption}
\usepackage{mwe}
\usepackage{multirow}
\usepackage{graphicx}
\usepackage[normalem]{ulem}
\useunder{\uline}{\ul}{}

\AtBeginDocument{%
  \providecommand\BibTeX{{%
    \normalfont B\kern-0.5em{\scshape i\kern-0.25em b}\kern-0.8em\TeX}}}

\setcopyright{none}
\copyrightyear{2018}
\acmYear{2018}
\acmDOI{XXXXXXX.XXXXXXX}

\acmConference[Conference acronym 'XX]{Make sure to enter the correct
  conference title from your rights confirmation emai}{June 03--05,
  2018}{Woodstock, NY}
\begin{document}

\title{Can ChatGPT Reproduce Human-Generated Labels? A Study of Social Computing Tasks}


\author{
    Yiming Zhu,\textsuperscript{\rm *}\quad
    Peixian Zhang,\quad
    Ehsan-Ul Haq,\quad
    Pan Hui,\textsuperscript{\rm *}\\
    Gareth Tyson\quad
}
\thanks{*~Yiming Zhu is also with The Hong Kong University of Science and Technology, and Pan Hui is also with The Hong Kong University of Science and Technology and University of Helsinki.}
\affiliation{
    \institution{The Hong Kong University of Science and Technology (Guangzhou)\\}
    \{yzhucd, euhaq\}@connect.ust.hk \quad pzhang041@connect.hkust-gz.edu.cn \quad
    \{panhui, gtyson\}@ust.hk
}

\renewcommand{\shortauthors}{Yiming Zhu, et al.}

\begin{abstract}

The release of ChatGPT has uncovered a range of possibilities whereby large language models (LLMs) can substitute human intelligence.
In this paper, we seek to understand whether ChatGPT has the potential to reproduce human-generated label annotations in social computing tasks. Such an achievement could significantly reduce the cost and complexity of social computing research.
As such, we use ChatGPT to re-label five seminal datasets covering stance detection (2x), sentiment analysis, hate speech, and bot detection.
Our results highlight
that ChatGPT \emph{does} have the potential to handle these data annotation tasks, although a number of challenges remain. 
ChatGPT obtains an average accuracy 0.609.
Performance is highest for the sentiment analysis dataset, with ChatGPT correctly annotating 64.9\% of tweets. Yet, we show that performance varies substantially across individual labels.
We believe this work can open up new lines of analysis and act as a basis for future research into the exploitation of ChatGPT for human annotation tasks. 
\end{abstract}





\maketitle

\section{Introduction} \label{sec:introduction}


The use of crowd-sourced human intelligence is common to annotate text~\cite{haq2022its,sorokin2009utility,akkaya2010amazon}. 
This human annotation generates ground truth data to train machine learning models for tasks such as stance detection~\cite{glandt2021stance}, hate speech detection~\cite{he2021racism}, sentiment analysis~\cite{rosenthal2019semeval}, and bot detection~\cite{fagni2021tweepfake}. While unsupervised methods are being introduced for classification tasks, such methods usually require large data samples~\cite{usama2019unsupervised, wang2020survey}. Thus, human annotation is still primarily used in social computing research.


Recently, the release of ChatGPT by OpenAI has uncovered a range of possibilities where large language models (LLMs) can help substitute human intelligence~\cite{guo2023close, bang2023multitask, zhang2023complete}. Several recent works compare the use of ChatGPT as compared to human and other automated methods~\cite{zhang2022would, sobania2023analysis, wang2023chatgpt}. For instance, researchers have investigated the use of ChatGPT for automatic bug fixing~\cite{sobania2023analysis}, misinformation detection~\cite{sallam2023chatgpt}, and even generating academic writings~\cite{aydin2022openai}.

In this paper, we explore the efficacy of using ChatGPT for five text-based social computing annotation tasks. We primarily seek to understand whether ChatGPT has the potential to reproduce these human-generated annotations. ChatGPT's annotations can highlight its usefulness against crowd-sourced annotations. 
To achieve this, we first use ChatGPT to label five seminal datasets on stance detection (2x), sentiment analysis, hate speech, and bot detection. 
We compare the ChatGPT's labels with the human assigned labels on those datasets. Our results show
that ChatGPT does have the potential to handle data annotation tasks.
Performance is highest for a sentiment analysis dataset, with ChatGPT correctly annotating 64.9\% of tweets. In contrast, performance is worst for the hate speech task, which correctly annotates 57.1\% of tweets.
For all five datasets, closer inspection reveals that performance varies substantially across individual labels.
For instance, while ChatGPT achieves a precision of only 0.353 in identifying tweets that contain anti-Asian hate speech, its precision increases to 0.791 when annotating tweets that contain counterarguments to such hate speech within the same task.



We believe this work can open up new lines of analysis and act as a basis for future research into the exploitation of ChatGPT for human annotation tasks.






\section{Related Work} \label{sec:related_work}


The release of ChatGPT has raised varied attention from research communities. This has led to its use in a variety of academic fields, such as programming~\cite{castelvecchi2022chatgpt,sobania2023analysis}, medical science~\cite{biswas2023chatgpt, pavlik2023collaborating, kung2023performance}, education~\cite{zhai2022chatgpt, choi2023chatgpt}, and social computing~\cite{taecharungroj2023can, haensch2023seeing}. 

In the field of natural language processing (NLP), recent research has also investigated ChatGPT's potential to act as an annotator for textual data, proposing insights into different NLP tasks.
For example, studies have looked at using ChatGPT for annotating misinformation~\cite{bang2023multitask, sallam2023chatgpt} and hate speech~\cite{huang2023chatgpt}.
Huang \etal report that ChatGPT is able to correctly annotate 80\% of the implicit hateful tweets from the \texttt{LatentHatred} dataset~\cite{elsherief2021latent}.
In addition, the authors show that ChatGPT's explanations can reinforce human annotators' perception of the target text in explaining why tweets would be annotated as hateful or not. 
Our study also examines how well ChatGPT performs in annotating hate speech. However, we do not limit the annotation to a binary decision for whether tweets are hateful or not, and include a neutral label.
Note, existing literature has highlighted the importance of this, stating that tweets with neutral expressions can act as a defense against the spread of hateful content~\cite{mathew2019thou, schieb2016governing}.


Others researchers have used ChatGPT for performing stance detection~\cite{zhang2022would, aiyappa2023can}. 
Zhang \etal~\cite{zhang2022would} evaluate ChatGPT's performance on detecting political stance on two prevalent datasets, \texttt{SemEval-2016}~\cite{mohammad2016semeval} and \texttt{P-Stance}~\cite{li2021p}.
The authors report that ChatGPT can outperform most state-of-art stance detection models in zero-shot settings, suggesting ChatGPT's potential to handle stance annotation tests.

Major challenges still remain though.
Aiyappa \etal~\cite{aiyappa2023can} discuss that ChatGPT's performance varies across different model versions. The authors also point out that such variance is due to the possibility of data leakage, where past prompts are used for training the next ChatGPT generation.
Nonetheless, it remains unclear how well ChatGPT performs in annotating individuals' stances in a broader context, such as those who are in favor or against a particular issue.
Similarly, our analysis finds further challenges, \eg that ChatGPT's has a tendency to overestimate neutral stances.

In addition to the aforementioned themes, others have experimented with ChatGPT to annotate tasks such as genre identification~\cite{kuzman2023chatgpt}, topic identification~\cite{gilardi2023chatgpt}, sentiment analysis~\cite{bang2023multitask}, and fake news detection~\cite{bang2023multitask, hoes2023using}. 
However, most of this literature only focuses on a single annotation task. 
In contrast, we seek to perform a comparative analysis across different data annotation tasks.





\section{Methodology} \label{sec:method}

\subsection{Overview}

We follow a comparative approach to analyze the differences in human annotation and ChatGPT annotations by utilizing five different datasets. 
We select five annotation tasks (and associated datasets) that are commonly used in academic research: \one Stance Detection~\cite{kuccuk2020stance}, \two Hate Speech Detection~\cite{fortuna2018survey}, \three Sentiment Detection~\cite{giachanou2016like}, \four Bot Detection~\cite{alharbi2021social}, and \five Russo-Ukrainian Stance Detection~\cite{cremisini2019challenging, zhu2022reddit, haq2022twitter}. 

For these five datasets, we then attempt to recreate the human annotations using ChatGPT. In the following subsections, we describe our datasets (\S\ref{sec:dataset}), and our approach for performing ChatGPT annotation (\S\ref{sec:chatgpt_annotation}).


\subsection{Datasets}\label{sec:dataset}

\begin{table*}[]
\centering
\resizebox{.85\textwidth}{!}{%
\begin{tabular}{l|cccc}
\hline
\textbf{Annotation Task} &
  \textbf{\#Tweet} &
  \textbf{Human Labels} &
  \textbf{\#Tweet Labeled by ChatGPT} &
  \textbf{ChatGPT's Labels} \\ \hline
\textbf{Stance Detection} &
  3658 &
  \begin{tabular}[c]{@{}c@{}}In-favor (37.3\%)\\ Against (23.9\%)\\ Neither (38.8\%)\end{tabular} &
  3649 (99.8\%) &
  \begin{tabular}[c]{@{}c@{}}In-favor (45.3\%)\\ Against (33.2\%)\\ Neither (21.5\%)\end{tabular} \\ \hline
\textbf{Hate Speech} &
  2289 &
  \begin{tabular}[c]{@{}c@{}}Hate (18.7\%)\\ Counterspeech (22.6\%)\\ Neutral (58.7\%)\end{tabular} &
  2267 (99.0\%) &
  \begin{tabular}[c]{@{}c@{}}Hate (51.7\%)\\ Counterspeech (17.6\%)\\ Neutral (30.7\%)\end{tabular} \\ \hline
\textbf{Sentiment Analysis} &
  9895 &
  \begin{tabular}[c]{@{}c@{}}Positive (42.7\%)\\ Negative (13.5\%)\\ Neutral (43.8\%)\end{tabular} &
  9879 (99.8\%) &
  \begin{tabular}[c]{@{}c@{}}Positive (27.1\%)\\ Negative (14.6\%)\\ Neutral (58.4\%)\end{tabular} \\ \hline
\textbf{Bot Detection} &
  17571 &
  \begin{tabular}[c]{@{}c@{}}Bot (46.1\%)\\ Human (53.9\%)\end{tabular} &
  16816 (99.9\%) &
  \begin{tabular}[c]{@{}c@{}}Bot (10.9\%)\\ Human (89.1\%)\end{tabular} \\ \hline
\textbf{Russo-Ukrainian Sentiment} &
  2205 &
  \begin{tabular}[c]{@{}c@{}}pro\_Ukraine (42.5\%)\\ pro\_Russia (24.5\%)\\ Not sure (33.0\%)\end{tabular} &
  2110 (95.7\%) &
  \begin{tabular}[c]{@{}c@{}}pro\_Ukraine (16.3\%)\\ pro\_Russia (31.0\%)\\ Not sure (52.7\%)\end{tabular} \\ \hline
\end{tabular}%
}
\caption{Descriptive statistics of selected datasets and labeling results by ChatGPT. The \# means \enquote{the number of}. The column \enquote{\#Tweet Labeled by ChatGPT} presents the number of tweets annotated by ChatGPT with one definite label.}
\label{tab:datasets}
\end{table*}

We use five available Twitter datasets on five different annotation tasks.
Our dataset selection strategy is based on the following requirements: 
\one The datasets must be in English to avoid differences in language provision~\cite{sera2002language}, 
\two The datasets must be annotated by human annotators, as we wish to compare the human annotations with chatGPT.
We list our targeted annotation tasks and corresponding datasets below.








\pb{Dataset 1: Stance Detection.}
We select a multi-topic
stance detection dataset called \texttt{COVID-19-Stance}~\cite{glandt2021stance}. This dataset contains COVID-19 tweets related to four topics -- \enquote{\textit{Wearing a Face Mask.}}, \enquote{\textit{Anthony S. Fauci, M.D.}}, \enquote{\textit{Keeping Schools Closed}}, and \enquote{\textit{Stay at Home Orders}}. Each tweet item is manually
annotated through crowdsourcing. The stance on each topic can fall into one of the three labels -- \enquote{In-favor}, \enquote{Against}, and \enquote{Neither}.

For the ChatGPT annotation, we utilize the training subset of this dataset provided with stance annotations. Note, \texttt{COVID-19-Stance} only releases the ids of tweets, so we retrieve the tweets' text utilizing \texttt{snscrape}~\footnote{\url{https://github.com/JustAnotherArchivist/snscrape}} and obtain  3658 tweets with text still available.


\pb{Dataset 2: Hate Speech.}
We select an anti-Asian hate and counterspeech dataset called \texttt{COVID-HATE}~\cite{he2021racism}. This dataset utilizes a collection of keywords and hashtags to collect tweets expressing anti-Asian hate, and countering hate speeches to support Asian ethnicity amidst COVID-19. The annotation is performed by pre-trained human annotators and validated by an agreement test. Each tweet item is annotated with one of three labels -- \enquote{Hate}, \enquote{Counterspeech}, and \enquote{Neutral}. 
For ChatGPT annotation, we utilize a subset released with the annotations of \texttt{COVID-HATE} and contains 2,289 tweets.


\pb{Dataset 3: Sentiment Analysis.}
We select a widely cited sentiment analysis dataset from \texttt{SemEval-2017 Task 4}~\cite{rosenthal2019semeval}.
We choose the dataset from Subtask A, where tweets are labeled as positive or negative. Each tweet is annotated by at least five annotators, who must pass hidden tests for quality control. The annotation task is to decide sentiment orientation for a tweet -- \enquote{Positive}, \enquote{Neutral}, \enquote{Negative}. 
For the ChatGPT annotation, we utilize the subset released for \enquote{Subtasks A} with sentiment annotations. In total, this dataset contains 9895 tweets. Note, since the human annotator for Subtask A are not informed of specific topics, we treat ChatGPT in this task similarly.
Thus, we do not pass it any topic through the prompts.


\pb{Dataset 4: Bot Detection.}
We select a deepfake tweet dataset called \texttt{TweepFake}~\cite{fagni2021tweepfake}. Here, the term deepfake refers to AI-generated content that are potentially deceptive~\cite{westerlund2019emergence}. 
This dataset contains tweets posted by 23 bots and 17 human accounts verified by the dataset's authors. The bots accounts and their content are generated by AI models (\eg{GPT-2, RNN, LSTM, etc.}). Each tweet item is annotated according to the account type of its author: \enquote{Bot} or \enquote{Human}. We utilize ChatGPT to annotate 17,571 items from \texttt{TweepFake}.

\pb{Dataset 5: Russo-Ukrainian Sentiment.}
We also investigate how ChatGPT performs on timely and recent domains that have arisen since its launch ($23^{th}$ November, 2022).
Thus, we create a tweets dataset for stance detection regarding to debates on the Russo-Ukrainian Sentiment~\footnote{Our Russo-Ukrainian Sentiment dataset is available at: \url{https://drive.google.com/file/d/1LytqtoVfp477t2FD4S1FxvFSKxYExHnh/view?usp=share_link}}. The dataset collects by the conflict-related keywords using Twitter API in English from from February 24 and March 6, 2022. Each tweet item is annotated based on its stance towards the invasion (\eg{support for Russia or Ukraine}).
We manually annotate 2205 tweets from this dataset, including 539 tweets labeled as "pro\_Russia", 938 tweets labeled as "pro\_Ukraine", and 728 tweets labeled as "not sure". We utilize ChatGPT to annotate these 2205 tweets.

\subsection{ChatGPT Annotation}\label{sec:chatgpt_annotation}

\begin{table*}[]
\centering
\resizebox{0.85\textwidth}{!}{%
\begin{tabular}{c|l}
\hline
\textbf{Tweet} &
  \begin{tabular}[c]{@{}l@{}}@sgtfible @CristinaAlesci @Walmart If you catch COVID19 and expire because\\ you're too stupid to wear a mask or social distance, then yeah, you probably WON'T\\ shop there anymore.\#Dumbass \#COVIDIOTS \#TrumpVirus \#WearADamnMask \#StayAtHomeSaveLives\end{tabular} \\ \hline
\textbf{Prompt} &
  \begin{tabular}[c]{@{}l@{}}Classify a COVID-19 tweet's stance on stay\_at\_home\_orders into Against, Favor, or, None.\\ Tweet: @sgtfible @CristinaAlesci @Walmart If you catch COVID19 and expire because\\ you're too stupid to wear a mask or social distance, then yeah, you probably WON'T\\ shop there anymore.\#Dumbass \#COVIDIOTS \#TrumpVirus \#WearADamnMask \#StayAtHomeSaveLives\\ Stance:\\ Explanation:\end{tabular} \\ \hline
\textbf{\begin{tabular}[c]{@{}c@{}}ChatGPT's\\ Response\end{tabular}} &
  \begin{tabular}[c]{@{}l@{}}Favor. The tweet is strongly in favor of stay-at-home orders and wearing masks to prevent \\ the spread of COVID-19. The use of hashtags such as \#StayAtHomeSaveLives and \#WearADamnMask \\ further emphasize the author's stance. The derogatory comments directed towards individuals \\ who do not follow these measures indicate a strong belief in their effectiveness.\end{tabular} \\ \hline
\end{tabular}%
}
\caption{Examples of tweet, prompt and ChatGPT's response.}
\label{tab:examples}
\end{table*}

For each annotated dataset, we try to recreate the same annotations using ChatGPT.
We utilize the ChatGPT API,\footnote{\url{https://openai.com/blog/introducing-chatgpt-and-whisper-apis}} configured with GPT module gpt-3.5-turbo, to annotate each target dataset. To set up our prompt strategy, we rely on an official prompt example
for classification tasks from OpenAI\footnote{\url{https://platform.openai.com/docs/guides/completion/prompt-design}} (see Appendix~\ref{appendix:openai_example}).

In the official document, most prompts are imperative sentences starting with a verb. 
As such, we choose ``Give'' and ``Classify'', which are frequently used in annotation work.
We use these verbs to design our prompt. 
According to the official template, we find that starting a new row with a word describing the subject and object in the prompt is effective.
Thus, we follow this pattern, injecting the subject and objective of the annotation task here.
Another benefit of this template is that it tends to result in ChatGPT's responses injecting the annotation label in the first sentence.

Based on this template, we modify this example into a generalized prompt template applicable to our five distinctive annotation tasks.
The template can be adjusted to be applied for annotation in different tasks as shown as follows:
\begin{displayquote}
Give/Classify the tweet a label about \texttt{[Topic]} from/into \texttt{[label1, label2, or label3]}.\\
Tweet: \enquote{\texttt{[text]}}\\
Label/Class: \\
Explanation:
\end{displayquote}


\noindent
where \texttt{[Topic]} refers to the topic or background of the tweet; \texttt{[label1, label2, or label3]} refers to the set of candidate labels for ChatGPT to annotate the tweet; and \texttt{[text]} refers to the text content of the tweet. 
We also provide two plain-language indexes which have the effect of instructing ChatGPT to reply with two pieces of information:
Label/Class and Explanation.
Label/Class indicates that ChatGPT should respond with the label or class for annotation; whereas Explanation indicates that ChatGPT needs to provide supporting explanation for its annotation as well.



We then apply this template to generate a ChtaGPT prompt according to the dataset's original annotation strategy. 
For example, for the \texttt{COVID-HATE} dataset, tweets focus on COVID-19 
and are classified into three labels -- \textit{Neutral}, \textit{Counterspeech}, and \textit{Hate}. Accordingly, we inform ChatGPT of the tweets' topic as COVID-19 and ask it to classify corresponding tweets into the same three labels. 
Below is an example of such a  prompt, where the bold text refers to the content we edited within the generalized template: 
\begin{displayquote}
Classify the tweet about \textbf{COVID-19} into: \textbf{Hate, Counterspeech, or Neutral}. \\
Tweet: \enquote{\textbf{for the last f**king time.... CORONAVIRUS IS NO EXCUSE TO BE RACIST AGAINST ASIANS https://t.co/nBHTadCKzK}}\\
Class: \\
Explanation: 
\end{displayquote} 

Following this, we pass all samples to ChatGPT for annotation. We then process ChatGPT's responses to extract the labels for classification, as well as the explanation for ChatGPT's labeling.
Through manual analysis, we find that ChatGPT tends to put its label result in the first sentence.
To highlight this, Table~\ref{tab:examples} shows an example tweet, its corresponding prompt and ChatGPT's response.


To extract the label, we therefore first check the first sentence of each response and automatically check the number of distinctive labels mentioned in it.
If there is only one distinctive label, we consider it as ChatGPT's label result. If there are multiple distinct labels within the first sentence, we resort to manual checks.
Thus, we manually identify ChatGPT's label according to the full text of its responses. 

In all, we only encounter an average of 2.28\% ($SD=2.45\%$) tweets that require manually checks across all five datasets. 
Finally, we treat the remaining sentences in ChatGPT's response as the explanation of its labeling.
Note, in a small number (2.5\%) of cases, ChatGPT states there is not enough information for it to make a decision. For example, 

\begin{displayquote}
\textit{\enquote{There is not enough information to determine whether the author of the tweet is a bot or a human. The tweet itself is incomplete, leaving the meaning of the statement unclear.}}
\end{displayquote}

We emphasize that there are many ways in which our methodology could be expanded and refined. Our future work will involve exploring alternative forms of prompt formulation and response mining.






\section{Results and Analysis}\label{sec:analysis}

To evaluate the performance of ChatGPT, we compare ChatGPT's annotation against the original human annotations contained within each dataset.
We treat the original human annotations as the gold standard that ChatGPT must predict. As such, we treat ChatGPT as a prediction engine, which we can then evaluate using traditional classifier metrics.
We use weighted F1-score to evaluate ChatGPT's performance on data annotation. 
Given a dataset, a higher F1-score indicates that ChatGPT provides annotations that are \emph{more} similar to humans (\ie more accurate).


\subsection{Annotation Data Summary}

Table~\ref{tab:datasets} presents statistics for the ChatGPT annotation results. The annotated datasets contain 35618 tweets.
ChatGPT annotates 34721 (97.5\%) tweets, while only 897 (2.5\%) tweets are stated to be unclear or lacking enough information to assign a label to by ChatGPT. This confirms that ChatGPT can generate easily extractable annotation labels in most cases. 

\subsection{Results Summary}
\label{sec:manual_annotation}

\begin{table}[]
\centering
\resizebox{\columnwidth}{!}{%
\begin{tabular}{l|cccc}
\hline
\textbf{Annotation Task} & \textbf{Accuracy} & \textbf{w-Precision} & \textbf{w-Recall} & \textbf{w-F1-Score} \\ \hline
\textbf{Stance Detection}      & 0.612 & 0.647 & 0.612 & 0.604 \\
\textbf{Hate Speech}           & 0.571 & 0.720 & 0.571 & 0.581 \\
\textbf{Sentiment Analysis}    & \textbf{0.649} & 0.684 & \textbf{0.650} & \textbf{0.646} \\
\textbf{Bot Detection}         & 0.639 & \textbf{0.761} & 0.639 & 0.572 \\
\textbf{Russo-Ukrainian Sentiment} & 0.573 & 0.642 & 0.573 & 0.551 \\ \hline
\end{tabular}%
}
\caption{Descriptive measurement for overall ChatGPT's performance in each annotation task. The prefix \enquote{w-} notes that the measurements' calculations are weighted by the number of true predictions for each label. The bolden numbers refer to the highest values for each measurement among the five annotation tasks.
}
\label{tab:measure_overall}
\end{table}

\begin{table}[]
\centering
\resizebox{\columnwidth}{!}{%
\begin{tabular}{l|lccc}
\hline
\textbf{Annotation Task}                        & \textbf{Label} & \textbf{Precision} & \textbf{Recall} & \textbf{F1-Score} \\ \hline
\multirow{3}{*}{\textbf{Stance Detection}}   & In-favor      & 0.591 & 0.719 & \textbf{0.649} \\
                                             & Against       & 0.540 & 0.747 & 0.627 \\
                                             & Neither       & 0.767 & 0.426 & 0.548 \\ \hline
\multirow{3}{*}{\textbf{Hate Speech}}        & Hate          & 0.353 & 0.969 & 0.518 \\
                                             & Counterspeech & 0.791 & 0.610 & \textbf{0.689} \\
                                             & Neutral       & 0.812 & 0.427 & 0.560 \\ \hline
\multirow{3}{*}{\textbf{Sentiment Analysis}} & Positive      & 0.813 & 0.514 & 0.630 \\
                                             & Negative      & 0.575 & 0.623 & 0.598 \\
                                             & Neutral       & 0.593 & 0.790 & \textbf{0.677} \\ \hline
\multirow{2}{*}{\textbf{Bot Detection}}      & Bot           & 0.951 & 0.226 & 0.364 \\
                                             & Human         & 0.601 & 0.990 & \textbf{0.748} \\ \hline
\multirow{3}{*}{\textbf{Russo-Ukrainian Sentiment}} & Pro-Ukraine       & 0.769              & 0.297           & 0.429             \\
                                             & Pro-Russia    & 0.671 & 0.827 & \textbf{0.733} \\
                                             & Not Sure      & 0.455 & 0.741 & 0.562 \\ \hline
\end{tabular}%
}
\caption{
Descriptive measurement for detailing labels. A bold number notes the highest F1-score in an annotation task.
}
\label{tab:my-table}
\end{table}

\begin{table}[]
\centering
\resizebox{\columnwidth}{!}{%
\begin{tabular}{l|cccc}
\hline
\textbf{Topic} & \textbf{Accuracy} & \textbf{w-Precision} & \textbf{w-Recall} & \textbf{w-F1-Score} \\ \hline
\textbf{Wearing a Face Mask.}    & 0.743 & 0.714 & 0.743 & 0.698 \\
\textbf{Anthony S. Fauci, M.D.}  & 0.565 & 0.577 & 0.565 & 0.527 \\
\textbf{Keeping Schools Closed.} & 0.451 & 0.493 & 0.451 & 0.465 \\
\textbf{Stay at Home Orders}     & 0.657 & 0.778 & 0.657 & 0.689 \\ \hline
\end{tabular}%
}
\caption{Descriptive measurement for overall ChatGPT's performance among different topics in COVID stance detection.}
\label{tab:measure_target}
\end{table}

Table~\ref{tab:measure_overall} presents the precision, recall rate, and F1-score for ChatGPT's predictions for each dataset. 
For the five annotation tasks, ChatGPT achieves an average accuracy of 0.609 ($SD=0.032$)
\textit{as a data annotator, ChatGPT has the potential to generate some annotations similar to humans.} However, its capability is still limited (weighted F1-score $<0.65$ for all five datasets).

To explore how ChatGPT performs on different labels in each annotation task,
Figure~\ref{fig:cm} presents the confusion matrices for the five annotation tasks. For each matrix, the y-axis refers to the ground truth human labels, and the x-axis refers to ChatGPT's labels.
The value in a cell, row \textbf{\emph{i}} and column \textbf{\emph{j}}, presents the proportion of tweets with label \textbf{\emph{i}} that are correctly annotated with label \textbf{\emph{j}} by ChatGPT. 

For all annotation tasks, the proportions of correctly annotated tweets by ChatGPT vary per distinct label.
For example, Figure~\ref{fig:cm_hatespeech} presents the confusion matrix for the Hate Speech task. Here, we see that ChatGPT corrects labels 61\% of  \textit{Counterspeech} samples \vs 97\% for \textit{Hate} \vs just 43\% for \textit{Neutral}. In this case, ChatGPT is much more accurate at annotating tweets containing hate speech than those with neutral expressions or counterarguments.

Moreover, we find that ChatGPT's performance significantly varies across different labels in the Bot Detection and Russo-Ukrainian Sentiment.
ChatGPT peforms better on \enquote{Pro-Russia} labels (F1-score $=0.733$) than \enquote{Pro-Ukraine} (F1-score $=0.429$) and \enquote{Not Sure} (F1-score $=0.562$) labels. 
In Bot Detection, ChatGPT also performs much better on \enquote{Human} (F1-score $=0.748$) than \enquote{Bot} (F1-score $=0.364$).
These results suggest that \textit{for a given annotation task, ChatGPT's performance varies heavily across different labels.}

\subsection{Task Analysis and Implications}

Next, we dive into each annotation task to present implications according to ChatGPT's performance.



\pb{Sentiment Analysis.}
For Sentiment Analysis, ChatGPT's overall performance ranks \textit{first} out of our five annotation tasks by a weighted F1-score of 0.646.
ChatGPT correctly annotates 64.9\% tweets' sentiment labels. This is the highest accuracy among the annotation tasks. In addition, the highest weighted recall rate of 0.65 suggests ChatGPT's advantage in identifying correct labels for annotating tweets' sentiment.
Moreover, ChatGPT also achieves the highest weighted F1-score (0.646) for the task.

\begin{figure*}
    \centering
    \begin{subfigure}[b]{0.30\textwidth}
        \centering
        \includegraphics[width=\textwidth]{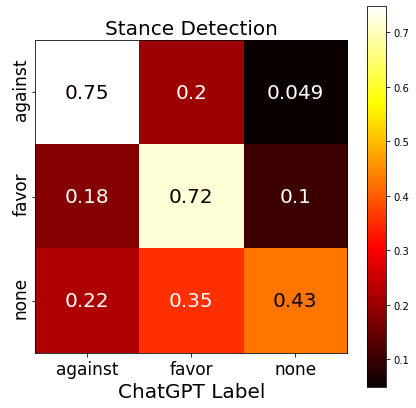}
        \caption{Stance Detection}%
        \label{fig:cm_stance}
    \end{subfigure}
    \hfill
    \begin{subfigure}[b]{0.30\textwidth}  
        \centering 
        \includegraphics[width=\textwidth]{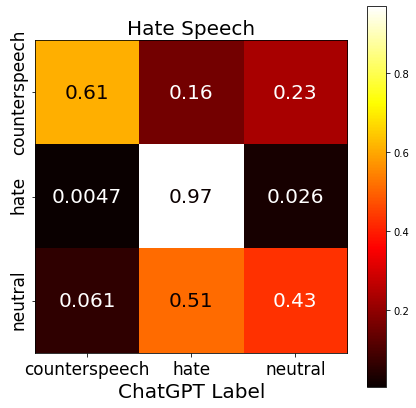}
        \caption{Hate Speech}%
        \label{fig:cm_hatespeech}
    \end{subfigure}
    \hfill
    \begin{subfigure}[b]{0.30\textwidth}   
        \centering 
        \includegraphics[width=\textwidth]{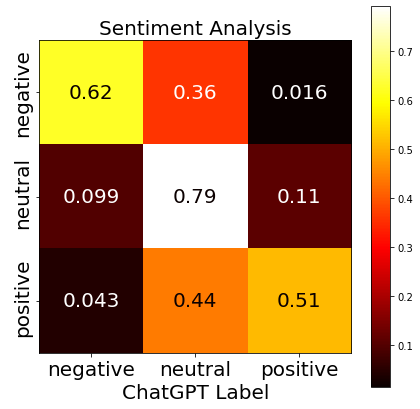}
        \caption{Sentiment Analysis}%
        \label{fig:cm_sentiment}
    \end{subfigure}
    \vskip\baselineskip
    \begin{subfigure}[b]{0.30\textwidth}   
        \centering 
        \includegraphics[width=\textwidth]{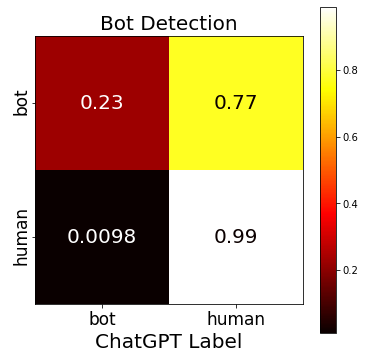}
        \caption{Bot Detection 
        }
        \label{fig:cm_bot}
    \end{subfigure}
    \hspace*{2cm}
    \begin{subfigure}[b]{0.30\textwidth}   
        \centering 
        \includegraphics[width=\textwidth]{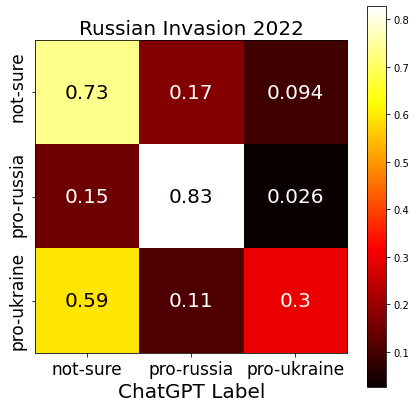}
        \caption{Russo-Ukrainian Sentiment}%
        \label{fig:cm_russian}
    \end{subfigure}
    \caption{
    The confusion matrices of ChatGPT's annotations for the five annotation tasks. The y-axis label for a given row shows the original label of tweets, and values in each cell show the percentage of those tweets classified in the corresponding x-axis label by ChatGPT.
    }
    
    \label{fig:cm}
\end{figure*}

Interestingly, as highlighted in the confusion matrix shown in Figure~\ref{fig:cm_sentiment}, we find that ChatGPT mis-annotates 36\% of negative tweets and 44\% of positive tweets with the \enquote{Neutral} label. 
This suggests that ChatGPT often mistakes high-polarity tweets as neutral but, importantly, not in the opposite polarity.
In conclusion, \textit{ChatGPT shows better overall performance on Sentiment Analysis than the other four annotation tasks.
Yet it also presents a tendency towards annotating the sentiment as neutral}.


\pb{Stance Detection.} For Stance Detection, ChatGPT's overall performance ranks \textit{second} out of our five annotation tasks by a weighted F1-score of 0.604.
ChatGPT correctly annotates 61.2\% of tweets' stances. In addition, ChatGPT's performance across different stances doesn't vary largely. Compared to \enquote{In-favor} (F1-score $=0.649$) and \enquote{Against} (F1-score $=0.627$), ChatGPT works not so well for \enquote{Neither} (F1-score $=0.548$). This is highlighted by the confusion matrix shown on Figure~\ref{fig:cm_stance}. 

35\% of non-stance tweets are mis-annotated with the \enquote{In-favor} label, and 22\% are mis-annotated with the \enquote{Against} label. \textit{This suggests that for stance detection, ChatGPT may have a tendency to overestimate certain stances, emphasizing the need for human moderation of labels.
}


\pb{Hate Speech.} 
For Hate Speech, ChatGPT's overall performance ranks \textit{third} out of our five annotation tasks by a weighted
F1-score of 0.581.
ChatGPT correctly annotates 57.1\% tweets.
ChatGPT achieves its highest precision for \enquote{Counterspeech} (precision $=0.791$) and \enquote{Neutral} (precision $=0.812$).
Yet, the \enquote{Hate} label's precision is much lower ($=0.353$). 
Moreover, ChatGPT attains a high recall rate of 0.969 and a low precision of 0.353 on \enquote{Hate}. This implies that, while ChatGPT can identify most of the hateful tweets (high recall), it tends to annotate many non-hateful tweets as hate speech incorrectly (low precision).

According to the confusion matrix in Figure~\ref{fig:cm_hatespeech}, we find that tweets with \enquote{Neutral} labels are more vulnerable to such miss-annotation, where 51\% neutral tweets have been annotated with \enquote{Hate} label by ChatGPT. 
To summarize, \textit{ChatGPT seems to  be better at annotating non-hateful content that includes tweets containing counterarguments, when compared against other tweets containing hateful or neutral expressions.
However, when it comes to predicting hate tweets, ChatGPT (using our prompt) often mis-annotate neutral tweets as hateful}. 

\pb{Bot Detection.}
For Sentiment Analysis, ChatGPT's overall performance ranks \textit{fourth} out of our five annotation tasks by a weighted
F1-score of 0.572. ChatGPT correctly annotates 63.9\% of tweets' labels.
Yet, we find that ChatGPT attains significantly different performance for the two labels, shown in Figure~\ref{fig:cm_bot}.
The \enquote{Human} label attains an F1-score of 0.748 compared to just 0.364 for the \enquote{Bot} label.
In addition, we find that, for the \enquote{Bot} label, ChatGPT reports a high precision of 0.951, with a very low recall rate only of 0.226. This means that, when ChatGPT annotates bot tweets, it is usually correct (high precision), but the same is not true for annotating a tweet as by a human (low recall). 

Such a result implies that, when encountering deep fake content, ChatGPT prefers a conservative annotation and does not risk annotating it as an incorrect \enquote{Bot} label if tweets are deceptive as human-generated. 
While ChatGPT correctly annotates 99\% of human tweets, it mis-annotates 77\% bot tweets with \enquote{Human} label.
In all, \textit{ChatGPT is conservative in annotating deepfake tweets as bot tweets, suggesting it has potential to act as a precise annotator to predict bot-generated content. Meanwhile, ChatGPT can be deceived to annotate deepfake tweets as human-generated}.

\pb{Russo-Ukrainian Sentiment.}
For Russo-Ukrainian Sentiment, ChatGPT's overall performance ranks \textit{fifth} out of our five annotation tasks by a weighted F1-score of 0.551 (the lowest among the five annotation tasks).
ChatGPT only correctly annotates 57.3\% tweets in the dataset. 
Interestingly, we notice that ChatGPT performs much better on annotating tweets expressing a pro-Russian stance. 
ChatGPT achieves an F1-score of 0.733 for \enquote{Pro-Russia}, compared to 0.429 for \enquote{Pro-Ukraine}, and 0.562 for \enquote{Not Sure}. 
We also observe an acceptable precision of 0.671 and a high recall rate of 0.827 for the \enquote{Pro-Russia} stance. 

In contrast, ChatGPT performs much worse on annotating tweets expressing a pro-Ukraine stance. Only 45.5\% of ChatGPT's predictions towards pro-Ukraine tweets are correct. Moreover, according to the confusion matrix shown in Figure~\ref{fig:cm_russian}, 59\% pro-Ukraine tweets are mis-annotated with a \enquote{Pro-Russia} label and 11\% are mis-annotated with \enquote{Not Sure} label. 
We have observed that \textit{when it comes to a recent domain such as Russo-Ukrainian Sentiment, ChatGPT's performance is potentially worse compared to a domain before ChatGPT's launch (November 30th, 2022), such as COVID-19 in our case. This suggests that ChatGPT's performance may be weaker in annotating data related to a new domain after its initial training.}
Compared to the \texttt{COVID-19-Stance} dataset detection, ChatGPT's performance is also more variable across different stances for the Russo-Ukrainian Sentiment. This implies that \textit{change of context can impact ChatGPT's performance on annotating tweets' stances}.





\section{Conclusion and Discussion}
\label{sec:discussion}

In this study, we have investigated ChatGPT's potential to act as a data annotator for different data annotation tasks. We have evaluated ChatGPT's performance on five datasets against the original human annotations. We use a standard prompt design that consists of a potential annotation list and a topic depending on the dataset.
Based on the analysis of the datasets, our work highlights ChatGPT's limitation for annotation tasks, as well as flagging areas that can help improve ChatGPT's annotation performance.

We find that ChatGPT's performance varies across the tweets' topics. For example, in the COVID stance detection task, ChatGPT's performance on Wearing a Face Mask (F1-score $=0.689$) varies significantly to the Keeping Schools Closed (F1-score $=0.465$) labels, as shown in Table~\ref{tab:measure_target}.
Our observations imply that ChatGPT performs better in cases where the annotation is based on knowledge about certain events, such as stance detection and sentiment analysis. However, in less objective cases such as bot detection, ChatGPT performs relatively weaker.
As a potential explanation, previous research has identified the power of other features in bot detection tasks (\eg social network information~\cite{beskow2018bot} and profile metadata~\cite{santia2019detecting}).
ChatGPT's limitation in incorporating such information in inferences may explain the low scores on such tasks.

We note that our work has a number of limitations, which could lead to interesting future work. Most notably, we only use a single prompt for annotation.
Prompt design is a major theme of future work, which we believe we yield better results. This first should involve specializing prompts for individual tasks.
We are keen to explore using ChatGPT to assist with prompt formulation.
Specifically, pur early experiments suggest that ChatGPT can be used to generate optimized prompts for individual tasks.
We also argue that developing new ways to embed secondary data (\eg context) in the prompts could improve results. This could be done via improved prompt formulation, or via multiple staged queries.
Our work could also be enhanced by combining more iterative and human-in-loop approaches~\cite{wang2021want} to interpret ChatGPT labels better and to clarify the answers. We hope that our work can act as a catalyst for further work in the area of automated text annotation.

\bibliographystyle{ACM-Reference-Format}
\bibliography{sample-base}

\appendix

\section{OpenAI's official prompt example for ChatGPT to conduct classification tasks.} \label{appendix:openai_example}

\begin{figure*}[]
\includegraphics[width=.85\textwidth]{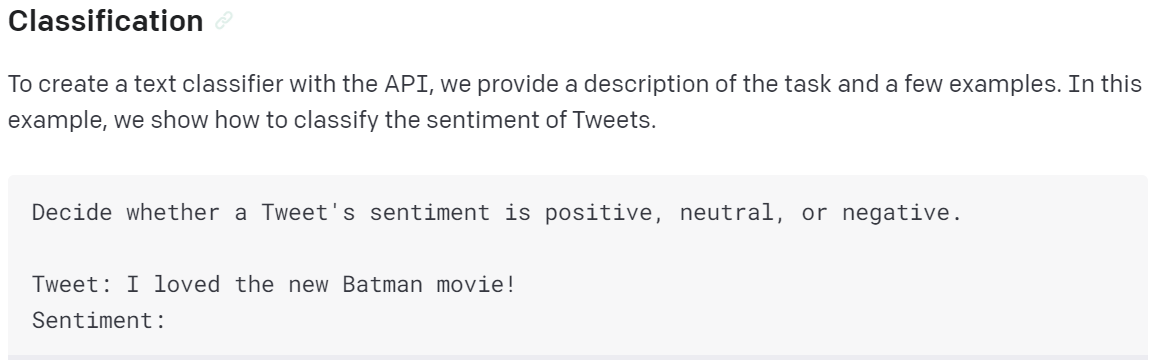}
\centering
\caption{OpenAI's official prompt example for ChatGPT to conduct classification tasks.}
\end{figure*}

\section{Examples of tweet, prompt and ChatGPT’s response.}\label{appendix:prompt}

\begin{table*}[]
\centering
\resizebox{\textwidth}{!}{%
\begin{tabular}{l|c|l}
\hline
\textbf{Task} &
  \textbf{Keys} &
  \textbf{Text Content} \\ \hline
\multirow{3}{*}{\textbf{\begin{tabular}[c]{@{}l@{}}Stance\\ Detection\end{tabular}}} &
  \textbf{Tweet} &
  \begin{tabular}[c]{@{}l@{}}@sgtfible @CristinaAlesci @Walmart If you catch COVID19 and expire because\\ you're too stupid to wear a mask or social distance, then yeah, you probably WON'T\\ shop there anymore.\#Dumbass \#COVIDIOTS \#TrumpVirus \#WearADamnMask \#StayAtHomeSaveLives\end{tabular} \\ \cline{2-3} 
 &
  \textbf{Prompt} &
  \begin{tabular}[c]{@{}l@{}}Classify a COVID-19 tweet's stance on stay\_at\_home\_orders into Against, Favor, or, None.\\ Tweet: ``@sgtfible @CristinaAlesci @Walmart If you catch COVID19 and expire because\\ you're too stupid to wear a mask or social distance, then yeah, you probably WON'T\\ shop there anymore.\#Dumbass \#COVIDIOTS \#TrumpVirus \#WearADamnMask \#StayAtHomeSaveLives''\\ Stance:\\ Explanation:\end{tabular} \\ \cline{2-3} 
 &
  \textbf{\begin{tabular}[c]{@{}c@{}}ChatGPT's\\ Response\end{tabular}} &
  \begin{tabular}[c]{@{}l@{}}Favor. The tweet is strongly in favor of stay-at-home orders and wearing masks to prevent \\ the spread of COVID-19. The use of hashtags such as \#StayAtHomeSaveLives and \#WearADamnMask \\ further emphasize the author's stance. The derogatory comments directed towards individuals \\ who do not follow these measures indicate a strong belief in their effectiveness.\end{tabular} \\ \hline
\multirow{3}{*}{\textbf{Hate Speech}} &
  \textbf{Tweet} &
  Are we still allowed to quote ancient Chinese proverbs, or is that racist? \#RacismIsAVirus \\ \cline{2-3} 
 &
  \textbf{Prompt} &
  \begin{tabular}[c]{@{}l@{}}Classify the tweet about COVID-19 into: Hate, Counterspeech, or Neutral.\\ Tweet: ``Are we still allowed to quote ancient Chinese proverbs, or is that racist? \#RacismIsAVirus''\\ Class:\\ Explanation:\end{tabular} \\ \cline{2-3} 
 &
  \textbf{\begin{tabular}[c]{@{}c@{}}ChatGPT's\\ Response\end{tabular}} &
  \begin{tabular}[c]{@{}l@{}}Counterspeech. \\ The tweet challenges the notion that quoting ancient Chinese proverbs is considered racist due to \\ the association of COVID-19 with China, and instead promotes the idea that racism is the real virus.\end{tabular} \\ \hline
\multirow{3}{*}{\textbf{\begin{tabular}[c]{@{}l@{}}Sentiment\\ Analysis\end{tabular}}} &
  \textbf{Tweet} &
  @thehuwdavies you think the Boro will beat Swansea? I'm not so sure, December/January is when we implode \\ \cline{2-3} 
 &
  \textbf{Prompt} &
  \begin{tabular}[c]{@{}l@{}}Give the tweet's sentiment a label from Positive, Negative, or Neutral.\\ Tweet: ``@thehuwdavies you think the Boro will beat Swansea? I'm not so sure, \\ December/January is when we implode''\\ Label: \\ Explanation:\end{tabular} \\ \cline{2-3} 
 &
  \textbf{\begin{tabular}[c]{@{}c@{}}ChatGPT's\\ Response\end{tabular}} &
  \begin{tabular}[c]{@{}l@{}}Neutral. The tweet is simply asking a question and expressing some uncertainty about the outcome,\\ without conveying particularly positive or negative sentiment.\end{tabular} \\ \hline
\multirow{3}{*}{\textbf{\begin{tabular}[c]{@{}l@{}}Bot \\ Detection\end{tabular}}} &
  \textbf{Tweet} &
  \begin{tabular}[c]{@{}l@{}}Had a wonderful interaction with leaders and MPs across party lines this evening. \\ We look forward to a productive Parliament session, where people-centric and \\ development oriented issues would be discussed. https://t.co/EsNze39bSA\end{tabular} \\ \cline{2-3} 
 &
  \textbf{Prompt} &
  \begin{tabular}[c]{@{}l@{}}Classify the author of the tweet into: Bot or Human.\\ Tweet: ``Had a wonderful interaction with leaders and MPs across party lines this evening. \\ We look forward to a productive Parliament session, where people-centric and \\ development oriented issues would be discussed. https://t.co/EsNze39bSA''\\ Author: \\ Explanation:\end{tabular} \\ \cline{2-3} 
 &
  \textbf{\begin{tabular}[c]{@{}c@{}}ChatGPT's\\ Response\end{tabular}} &
  \begin{tabular}[c]{@{}l@{}}The tweet shows a positive sentiment and uses natural language with no grammatical errors. \\ It also mentions specific events and issues, indicating that it came from a human who has \\ a better understanding of context and can convey more meaningful messages than bots.\end{tabular} \\ \hline
\multirow{3}{*}{\textbf{\begin{tabular}[c]{@{}l@{}}Rissian \\ Invasion\\ 2022\end{tabular}}} &
  \textbf{Tweet} &
  \begin{tabular}[c]{@{}l@{}}International Criminal Court : Stop Putin's War Crimes - Sign the Petition!\\ https://t.co/NyaFp6TTNj via @Change\end{tabular} \\ \cline{2-3} 
 &
  \textbf{Prompt} &
  \begin{tabular}[c]{@{}l@{}}Give the tweet about Russo-Ukrainian Sentiment a label from Pro-russia, Pro-ukraine, or Not Sure\\ Tweet: ``International Criminal Court : Stop Putin's War Crimes - Sign the Petition!\\ https://t.co/NyaFp6TTNj via @Change''\\ Label: \\ Explanation:\end{tabular} \\ \cline{2-3} 
 &
  \textbf{\begin{tabular}[c]{@{}c@{}}ChatGPT's\\ Response\end{tabular}} &
  \begin{tabular}[c]{@{}l@{}}Pro-Ukraine. The tweet is calling for the International Criminal Court to stop Putin's war crimes,\\ indicating an anti-Russian sentiment in favor of Ukraine.\end{tabular} \\ \hline
\end{tabular}%
}
\caption{Examples of tweet, prompt and ChatGPT’s response for our five annotation tasks.}
\label{tab:all_sample}
\end{table*}

\end{document}